\documentclass[10pt,twocolumn,letterpaper]{article}
\usepackage{cvpr}              

\usepackage[dvipsnames]{xcolor}

\usepackage{soul}
\usepackage{afterpage} 

\definecolor{cvprblue}{rgb}{0.21,0.49,0.74}
\usepackage[pagebackref,breaklinks,colorlinks,citecolor=cvprblue]{hyperref}
\usepackage{float}
\usepackage{subcaption}

\def\paperID{15} 
\def\confName{CVPR}
\def\confYear{2024}

\title{Shape-Preserving Generation of Food Images for Automatic Dietary Assessment}

\author{Guangzong Chen \qquad Zhi-Hong Mao\qquad Mingui Sun\qquad Kangni Liu \qquad Wenyan Jia \\
University of Pittsburgh \\
{\tt\small {\{guangzong, zhm4, drsun, connie.liu, wej6\}}@pitt.edu}
\and
}

\begin{document}
\maketitle
\section*{Abstract}

Traditional dietary assessment methods heavily rely on self-reporting, which is time-consuming and prone to bias. Recent advancements in Artificial Intelligence (AI) have revealed new possibilities for dietary assessment, particularly through analysis of food images. Recognizing foods and estimating food volumes from images are known as the key procedures for automatic dietary assessment. However, both procedures required large amounts of training images labeled with food names and volumes, which are currently unavailable. Alternatively, recent studies have indicated that training images can be artificially generated using Generative Adversarial Networks (GANs). Nonetheless, convenient generation of large amounts of food images with known volumes remain a challenge with the existing techniques. In this work, we present a simple GAN-based neural network architecture for conditional food image generation. The shapes of the food and container in the generated images closely resemble those in the reference input image. Our experiments demonstrate the realism of the generated images and shape-preserving capabilities of the proposed framework.

\section{Introduction}
Nutrition plays a pivotal role in maintaining health, influencing both our daily well-being and long-term health status. A balanced diet can foster overall wellness, while unhealthy eating habits can lead to a range of health problems, such as diabetes, heart disease, obesity, stroke, and certain types of cancers~\cite{Forouzanfar2015, Cecchini2010, Schatzkin2009, Willett2012}. Therefore, accurate dietary assessment is a critical component in keeping healthy and treating chronic diseases~\cite{Lieffers2012, Thompson2017}.

Traditional self-reported dietary assessment methods include 24-hour dietary recall (24HR), dietary records, and food frequency questionnaires (FFQ)~\cite{Thompson2017, Shim2014, Gibson2005, Ortega2015, Baranowski2012}. All these methods necessitate individuals to report their food consumption, detailing the type and/or volume of food consumed. However, such a process can be time-consuming, cumbersome, and biased, since it relies heavily on self-reporting. Individuals tend to report healthier food choices while neglecting unhealthy items. The reliance on self-reporting introduces potential inaccuracies in capturing a comprehensive and precise picture of an individual's dietary habits~\cite{Slimani2011, Moshfegh2008}. 

Food images can be conveniently acquired by wearable devices or smartphones, and thus image-assisted dietary assessment has attracted research interest and been extensively investigated. The integration of artificial intelligence (AI), especially deep learning networks, in analyzing food images has markedly advanced the automation of dietary assessment~\cite{Konstantakopoulos2024, Sultana2023, Dalakleidi2022}. Developing AI algorithms for dietary assessment requires a substantial collection of labeled images covering a wide range of food types and volumes for effective training. Manual labeling becomes necessary to fulfill this demand, which is a laborious and time-intensive task. While several datasets with large amounts of food images are currently available, there is still a need for training images to recognize foods in specific countries and regions~\cite{bossard14, Chen2016a, Thames2021, Kawano2014b}. Additionally, many existing food image datasets lack annotations for food volume, making them unsuitable for dietary assessment. 
To address these challenges, generative models have been proposed to synthesize images, thereby augmenting training image datasets. Generative Adversarial Networks (GANs) have demonstrated a powerful ability in several areas of image generation, such as super-resolution image generation, image inpainting, and image semantic editing. Although training images can be artificially generated using GANs, it is difficult for existing techniques to maintain both food shapes and image quality simultaneously, significantly affecting the accuracy of dietary assessment. In response, we propose a simple GAN-based network architecture. Our experiments indicate that the new form of GAN can not only generate realistic food images but also preserve the food shape in the reference image. The proposed approach can significantly enhance the performance of AI-based dietary assessment systems by generating training images for both food recognition and volume estimation. It will thus offer an effective, efficient, and scalable solution to overcome the current limitations in automatic dietary assessment.

The major contributions of this paper are twofold. First, we present a straightforward GAN architecture for realistic food image translation. Second, we demonstrate that, in the generated images, it is convenient to control food categories and preserve food shapes using style and category variables.

\section{Related Works}
\subsection{Automatic Dietary Assessment}

In the field of automated image-based dietary assessment, identifying and quantifying food nutrition, particularly recognizing their types and estimating volume, poses considerable challenges due to the complicated visual characteristics of various foods and the absence of reference scales in images~\cite{Lo2020, Konstantakopoulos2024, Raju2021, Amoutzopoulos2020}. Traditional food recognition relies on the extraction of image features, such as the scale-invariant feature transform (SIFT) and the histogram of oriented gradients (HOG), followed by classification using a classifier like the support vector machine (SVM)~\cite{Chen2012, He2014, Matsuda2012, bossard14, Konstantakopoulos2024}. However, the classification accuracy is low, and the algorithm is difficult to develop~\cite{Chen2012, He2014, bossard14}. Recently, with the rapid development of deep learning, deep networks and strategies such as fine-tuning and transfer learning have been effectively employed in the analysis of food images, leading to unprecedented levels of accuracy of food recognition~\cite{Konstantakopoulos2024,tan2019efficientnet, Arslan2022, Martinel2018, Min2019, Zhao2020, Hassannejad2016, Rajesh2022, Yanai2015}. For training and evaluating the deep network, food image datasets, such as Food-101, and UEC-Food, have been constructed using online sources (e.g., Google Images, Flickr) or collected images for different types of cuisines or specifically controlled environments~\cite{Konstantakopoulos2024, bossard14, Ciocca2017, Min2020, Okamoto2021, Sonmez2023, Kawano2014a, Ege2019, Wu2021}. Currently, the application of state-of-the-art methodologies to these datasets has achieved an impressive accuracy rate~\cite{tan2019efficientnet}. For example, the EfficientNet-B7 network achieves 93\% accuracy in the Food-101 dataset~(\cite{tan2019efficientnet}) and the ensemble method averaging the predictions of ResNeXt and DenseNet models reaches 90.02\% in the UEC-Food100 dataset~\cite{Konstantakopoulos2024}.

The challenge in calculating the volume of food from a single image is primarily attributed to the absence of three-dimensional (3D) information inherent in a two-dimensional (2D) image~\cite{Lo2020, Tahir2021, Konstantakopoulos2024, Tay2020}. Previous studies mostly rely on model-based techniques~\cite{Chen2013, Chae2011, Jia2014, Fang2015, Smith2022}. After a calibration procedure using a reference object with a known size (e.g., a checkerboard, credit card) to determine the camera's location and orientation, a pre-defined shape model is chosen for each food item to match the contour of the food and estimate its volume. However, this procedure is labor-demanding in most cases since manual operations are required, and estimating the volume of irregularly shaped food can be challenging~\cite{Jia2014, Fang2015, Smith2022}. Recently deep neural networks are expected to automatically learn the scale information of a 2D image from the global cures in the image and use it for volume estimation. Yang et al. propose a novel human-mimetic AI system to virtually gauge the volume of food using a set of internal reference volumes, mimicking the thinking of dietitians who mentally use a standard measuring tool (e.g., cup) as a reference~\cite{Yang2021}. Several studies employed convolutional neural networks (CNNs) to estimate a depth map or 3D shape (represented by voxels) corresponding to the input food image and obtain volumetric information~\cite{Christ2017, Myers2015, Fang2019, Shao2023}. In most of these studies, the training images were created by the research group themselves, either manually labeled~\cite{Yang2021, Fang2019} or captured with a depth sensor~\cite{Christ2017, Myers2015}. However, obtaining training images with labeled food volume/calorie or depth map is a tedious task. Thus, large-scale food image databases with known volume/nutrient information have not yet been developed.

\subsection {Food Image Generation}

It is well known that the quantity and quality of images in the training set play a critical role in the performance and generalization ability of deep networks. Therefore, data augmentation techniques (such as random crop, rotation, translation, flip, and rescaling) have been proposed to expand training datasets. To further increase the diversity of images, GANs have proven to be invaluable tools~\cite{Goodfellow2014, Luo2021a, Wang2020}. GANs introduce a novel approach to image generation by training a generator network to produce realistic images that are indistinguishable from real ones, while a discriminator network learns to differentiate between real and generated images.

Several GAN-based structures have been proposed to generate images from a list of ingredients/recipes or reference images~\cite{Ito2018, Zhu2020, Pan2020, Sugiyama2021}. The Multi-ingredient Pizza Generator (MPG) is a conditional GAN framework based on StyleGAN2 designed to generate pizza images with desired ingredients\cite{Han2020a}. CookGAN combines an attention-based recipe association model and StackGAN to generate meal images from ingredients~\cite{Zhu2020}. ChefGAN, RDE-GAN, and other related works integrate an image-recipe embedding module into GANs structure to synthesize dish images ~\cite{Pan2020, Sugiyama2021,  Wang2019b}.  

RamenGAN uses a conditional GAN to generate ramen images after training with a ramen image dataset~\cite{Ito2018}. arCycleGAN introduces the mechanism of attribute registration into CycleGAN to transfer the freshness styles from the style-offering images to the input images~\cite{Chen2021}. DuDGAN improves class-conditional GANs to control the output image using an additional classifier trained with a diffusion-based noise injection process~\cite{Yeom2023}. TransferI2I explores several novel techniques to implement image-to-image translation with limited data labeled data for two-class and multi-class translation tasks~\cite{Wang2021}. TUNIT is a truly unsupervised image-to-image translation model that simultaneously learns to separate image domains and translates input images into the estimated domains~\cite{Baek2021}. Besides the GAN structure, diffusion models have also been introduced recently to generate food images~\cite{Han2023, Markham2023}. 

Although promising results have been demonstrated in these studies, currently, recipe-image pairs are only available in the Recipe1M+ dataset~\cite{Marin2021}, and the volumes of the foods in the images generated from the recipe are unknown since they cannot be controlled. Thus in this work, we focus on image-to-image translation approaches designed to produce food images with the volumetric information. We aim to estimate the food volume from a single image, which is a projection of food in 3D. Therefore, the volume of the food is preserved, if the shape and depth map of the projection are unchanged. In our case, we assume that a small set of training images with known volumes exists but its size is insufficient for training. Our goal is to increase the size of this small training set by including new image samples which are created by replacing the foods in the existing images with numerous other foods. As a result, the combined set of images, which may be very large, can then be used to train deep neural networks for volume estimation.

In addition to preserving contours, maintaining the shapes of food containers is equally important since containers serve as references for estimating food volumes. In doing so, the realism of the generated images is enhanced. While CycleGAN, among various GAN structures, can maintain shapes in the generated images, retraining is necessary for each new class of images and this procedure is inefficient. It requires extra training to solve container distortion by introducing a discriminator to identify whether a dish plate observed exhibits a correct round shape, and the results are often not satisfactory~\cite {Ito2018}. A mask-based image synthesis network has been proposed to ensure a reasonable plate shape in generated images, but images with segmented plate regions are necessary~\cite{Honbu2022}. We propose a simple network to generate diverse, high-quality images while preserving the shapes of both the food and the container of a given dish in the reference image.

\color{black}
\section{Methods} 

\begin{figure*}[tp]
    \centering
    \includegraphics[height=5.3cm]{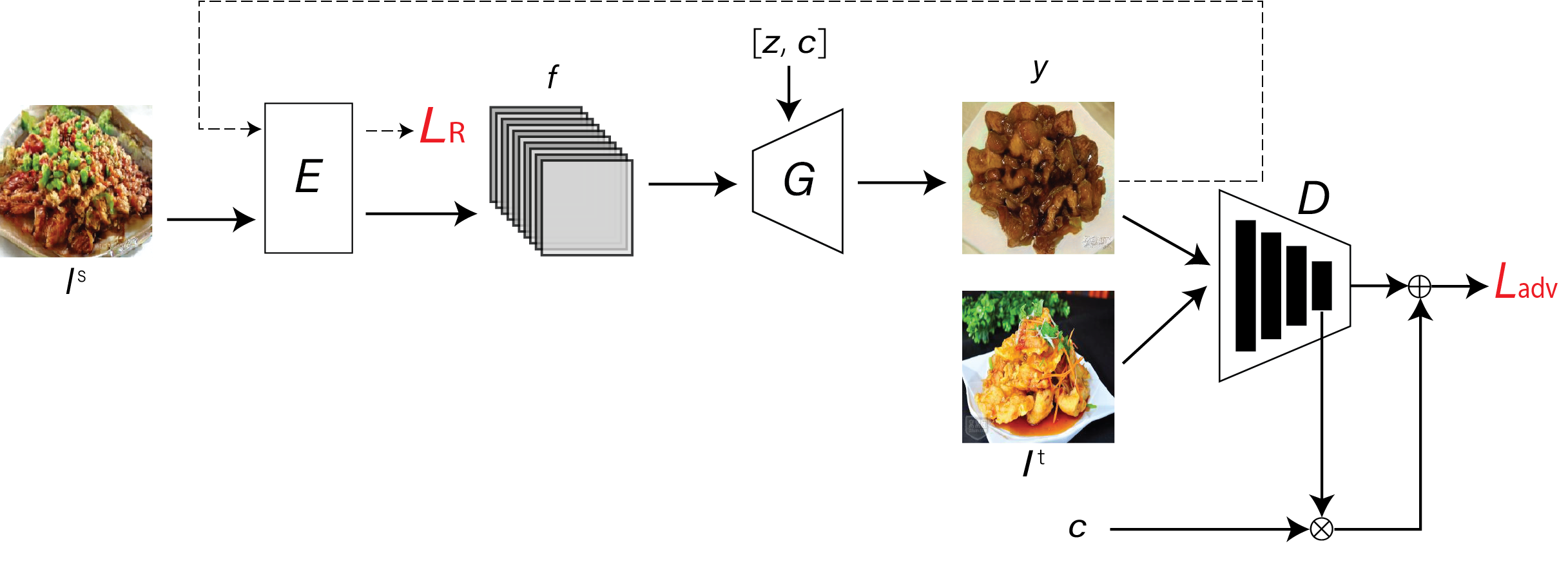}
    \caption{Our network architecture includes three major components, encoder $E$, generator $G$, and discriminator $D$. The encoder produces shape-related features $f$ from the image $I^{\text{s}}$. The generator takes features $f$, latent variable $z$, and category label $c$ as conditional inputs and create output image $y$. The discriminator is used to evaluate the realism of the output image. Loss functions $L_{\text{adv}}$ and $L_{\text{R}}$ are used for training the network.}
    \label{img:overview}
\end{figure*}

We develop a neural network architecture for image generation with specific object constraints. Our goal is to generate an image that retains the same object shape as the given reference image, while the textures are determined by a latent variable.
This variable enables the creation of diverse food images with identical shapes. The ``shape'' in this context includes both the shapes of the food and food container. We also use a category label as a conditional variable to control the object category of the generated image. The architecture of our network is illustrated in Fig.~\ref{img:overview}.

The network includes three parts. The first is an encoder, which compresses the input image into features. The second part is a generator, which takes the features and a latent variable as inputs, generating an image. The third part is a discriminator, which is used to distinguish between the real and generated images.

Compared with regular GANs, our proposed network architecture includes a shape encoder. The encoder is necessary for shape learning. Our model is remarkably compact, comprising only a single generator and a discriminator without the need for additional components. Given one shape image, multiple food images can be generated from our model. The generated images can be utilized to estimate the food volume in future research. The type of food can be specified through conditions, allowing the images to be used for training a food recognition network. As is widely acknowledged, defining image attributes, especially shapes, is challenging. Shape information cannot be accurately represented by just a few feature variables, making it impractical to use a classifier for defining shape features. Alternatively, we employ the encoder to extract shape information directly from images.  

Two datasets are used for training. The first image dataset, $\mathcal{I}^{\text{s}}$, is used as food shape references, where $\mathcal{I}^{\text{s}}$ equals $\{I^{\text{s}}_i\,|\, i = 1,\dots, N\}$,  $I^{\text{s}}_i \in \mathbb{R}^{ H \times W \times 3}$, $H$ and $W$ are the height and width of the images, $3$ is the number of channels of an RGB image, and $N$ is the total number of images.
The second image dataset, $\mathcal{I}_\text{t}$, is used to provide the food texture information, 
where $\mathcal{I}^{\text{t}}$ equals $\{I^{\text{t}}_i\,|\, i = 1,\dots, M\}$, $I^{\text{t}}_i \in \mathbb{R}^{ H \times W \times 3}$, and $M$ the size of the second dataset.
The ``textures'' mainly encompass various aspects such as the color of the material, grain size, condensed state, and other detailed characteristics of the food. This dataset is provided to the discriminator, ${D}$, to train the network. We want to apply $\mathcal{I}^{\text{s}}$ to facilitate the network to generate images with the same shapes as the images in $\mathcal{I}^{\text{s}}$ while maintaining the textures from $\mathcal{I}^{\text{t}}$.

\subsection{Functions of Network Components}

\textbf{Encoder.\ }
The input images, $I^{\text{s}}$, are compressed by the encoder to extract essential features. These features mainly contain the topological information of an image. The encoder also helps to reduce the resolution of the shape images, leading to a more compact network structure.

\begin{figure}[tp]
    \centering
    \includegraphics[width=0.7\linewidth]{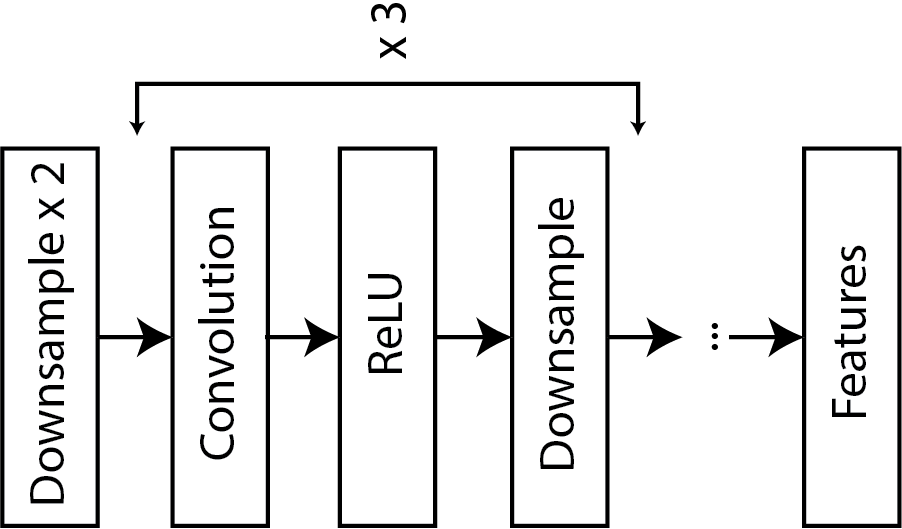}
    \caption{The network structure of the encoder.}
    \label{img:encoder}
\end{figure}

The encoder structure is shown in Fig.~\ref{img:encoder}. It consists of downsampling layers and convolutional layers. As the shape features of an image often encode global and topological information, which is of ``low-frequency'' nature, we apply two downsampling layers to reduce the resolution. This is followed by a stack of three convolutional network blocks, each containing a convolutional layer, a ReLU activation layer, and a downsampling layer.  

We use ${E}$ to represent the encoder model. The encoder takes an image $I^{\text{s}}$ from the shape dataset as input and outputs a feature vector, which is set of feature maps $f = {E}(I^{\text{s}}) \in \mathbb{R}^{H' \times W' \times C'}$, where $H'$ and $W'$ are the height and width of a feature map and $C'$ is the number of feature maps. We set $H' = W' = 16$ and $C' = 128$ in the experiment.

In the subsequent stages, the feature vector $f$ provides constraints for generated images, playing a pivotal role in defining the overall shapes of the objects within the generated images.

\noindent\textbf{Generator. \ }
The generator's primary role is to create images that adhere to the constraints derived from the shape features. The generator is designed with three inputs: the shape feature vector $f$, extracted by the encoder, the latent variable $z$, and the category label $c$. The shape feature vector $f$ primarily determines the shape of the object. The latent variable $z$, which is sampled from a Gaussian distribution, influences the texture of the generated image. The category label $c$ determines the image's class. The separated input ports of $f$, $z$, and $c$ are essential for isolating the shape, texture, and category of the generated image.

Let ${G}$ denote the generator model of the proposed network. 
The output image $y = {G}(f, z, c)$ is determined by the shape feature vector $f = E(I^{\text{s}})$ (output of the encoder), $z$ (latent variable), and $c$ the category. The dimension of $y$ is the same as that of $I^{\text{s}}$.

\noindent\textbf{Discriminator.\ }
For the discriminator $D$, a conditional discriminator with class embedding is used, which is similar to BigGAN~\cite{Brock2018}. The discriminator takes the generated image $y$ and an image sample $I^{\text{t}}$ from the texture dataset $\mathcal{I}^{\text{t}}$ as inputs and provides metrics for realism evaluation, which are $D(y, c_y) \in \mathbb{R}^{1}$, and $D(I^{\text{t}}, c_{I^{\text{t}}}) \in \mathbb{R}^{1}$.
The learning process for the discriminator is to detect the differences between the generated images and real images.

\subsection{Network Training}

Training is performed in two alternating stages. One stage is to train the encoder and generator, and the other is to train the discriminator. Different loss functions are applied in different stages. We apply a reconstruction loss and GAN loss when training the encoder and generator. The reconstruction loss ensures the same shape features are shared by the input and generated images. The GAN loss function and $R_1$~\cite{Mescheder2018} regularization are applied to train the discriminator.
We adopt the GAN loss from~\cite{Goodfellow2014} given by
\begin{equation}
\begin{aligned}
{L}_{\text{adv}} =&\, \mathbb{E}_{I^{\text{t}}} [\log(D(I^{\text{t}}, c_{I^{\text{t}}}))]  \\ +\,& \mathbb{E}_{I^{\text{s}}, z, c } [1 - \log (D({G}(E(I^{\text{s}}), z, c), c))]
\end{aligned}
\label{loss_gan}
\end{equation}
where $\mathbb{E}$ means expectation.
The $L_1$ loss is used as our reconstruction loss:
\begin{equation}
{L}_{\text{R}} = \mathbb{E}_{I^{\text{s}}, z, c } [\| E(I^{\text{s}}) - E({G}(E(I^{\text{s}}), z, c)) \|_1].
\label{loss_gen}
\end{equation}
Our learning problem is to solve
\begin{equation}
    \min_{E,\ G} \max_D \ \ {L}_{\text{adv}} + \lambda {L}_{\text{R}} 
\end{equation}
where $\lambda$ is a hyper-parameter indicating the relative weight of the reconstruction loss with respect to the GAN loss.
\color{black}

\section{Experiments}

\subsection{Realism Evaluation of Generated Food Images}
\label{exp1}

The primary objective of the first experiment is to demonstrate the capability of our method to generate realistic food images.
The quality of the generated images was quantitatively evaluated using the FID (Frechet Inception Distance)~\cite{Bynagari2019}, a widely used metric for assessing the fidelity of the generated images. To validate the effectiveness of our approach, we also conducted a comparative analysis with StyleGAN3~\cite{Karras2021}.
\begin{figure*}[tp]
    \centering
    \includegraphics[height=6.5cm]{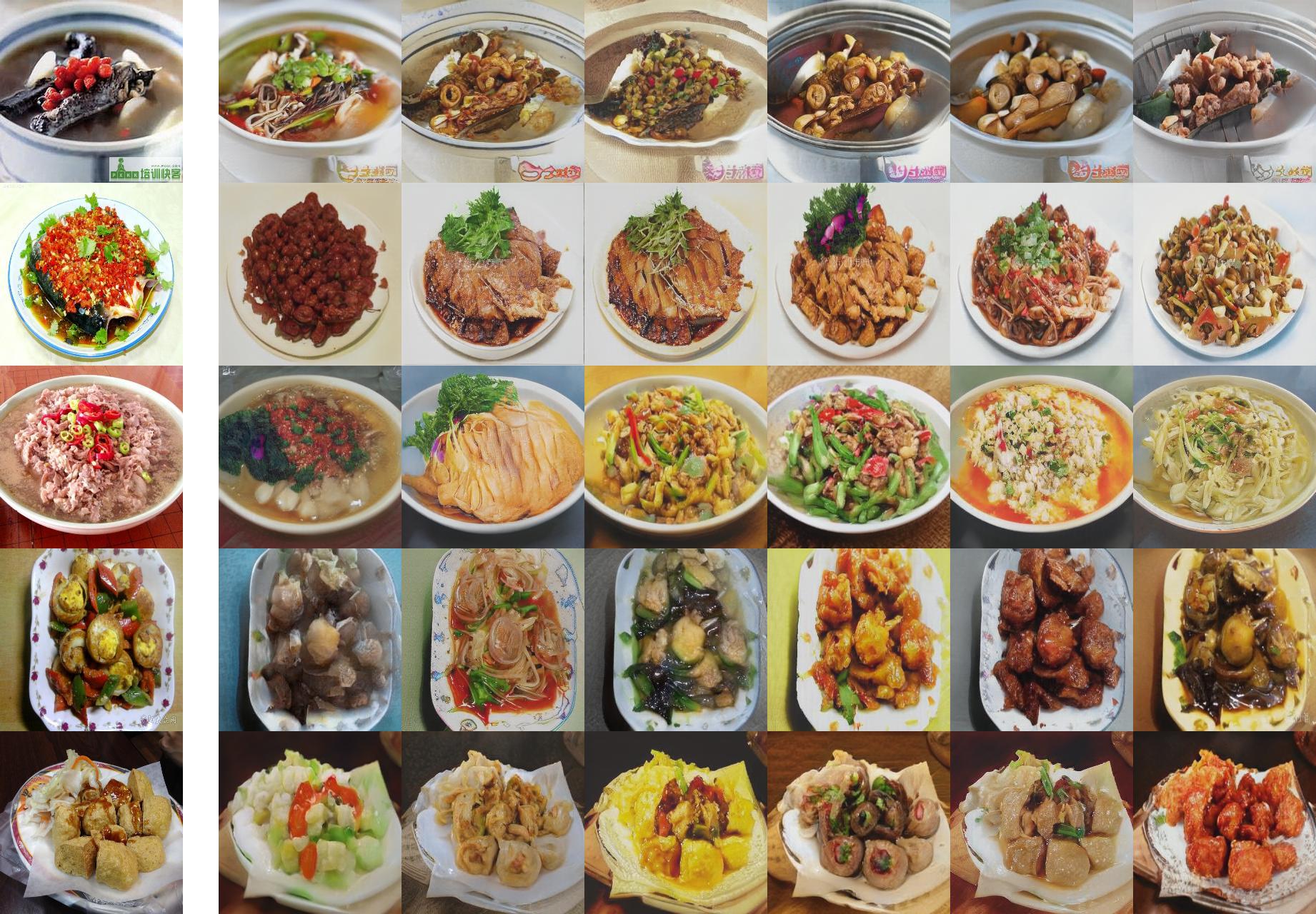}
    \vspace{-0.5em}
    \caption[short]{Image examples generated by our network using VireoFood-172 dataset: The first column shows the original input images, and subsequent columns display images created by varying the latent variable $z$ while keeping the corresponding input image from the first column fixed.}
    \label{img:food_gen}
\end{figure*}

\noindent\textbf{Datasets.\ }
A Chinese food image dataset VireoFood-172~\cite{Chen2016a} and a Western food image dataset Food-101~\cite{bossard14} were used to evaluate the performance of our approach. The VireoFood-172 dataset encompasses 172 distinct classes of Chinese food, with each class featuring between 300 and 1000 images. In total, the VireoFood-172 dataset comprises 110,241 images. Most images in this dataset contain a food item in a container (e.g., plate, bowl). Whether the shape of the container can be preserved was also studied in this experiment. The Food-101 dataset contains 101,000 images of 101 different food classes. 
As this dataset has been used by other researchers for food image generation, we also evaluated our approach with the Food-101 dataset for comparison with other approaches.
Before training the network, the images in both datasets were resized to $256 \times 256$ pixels to improve computational efficiency.

\noindent\textbf{Evaluation Metric.\ }
We used FID as the metric to evaluate the quality of the generated images. FID measures the discrepancy between the features of the generated and real images. These features are extracted using the Inception network~\cite{Szegedy2015}. The computation of FID involves comparing the distributions of these features as derived from the Inception network. A lower FID value signifies higher realism in the generated images. At the extreme, a zero value of FID indicates a perfect match in the distribution of the generated and real data, implying that the generated images are indistinguishable from the real. 

\noindent\textbf{Results.\ }
Fig.~\ref{img:food_gen} displays some image examples created by our network, which was trained by the VireoFood-172 dataset. We selected five random images as inputs, with the first column showing these inputs and the subsequent columns presenting the outputs generated by our network. 
These outputs were created by combining the same input image in each row with different style variables, indicated by $z$. This resulted in a notable change in textures, yielding highly realistic food visuals. 

\begin{figure*}[tp]
\centering
    \begin{subfigure}[t]{0.45\textwidth}
        \centering
        \includegraphics[height=4.5cm]{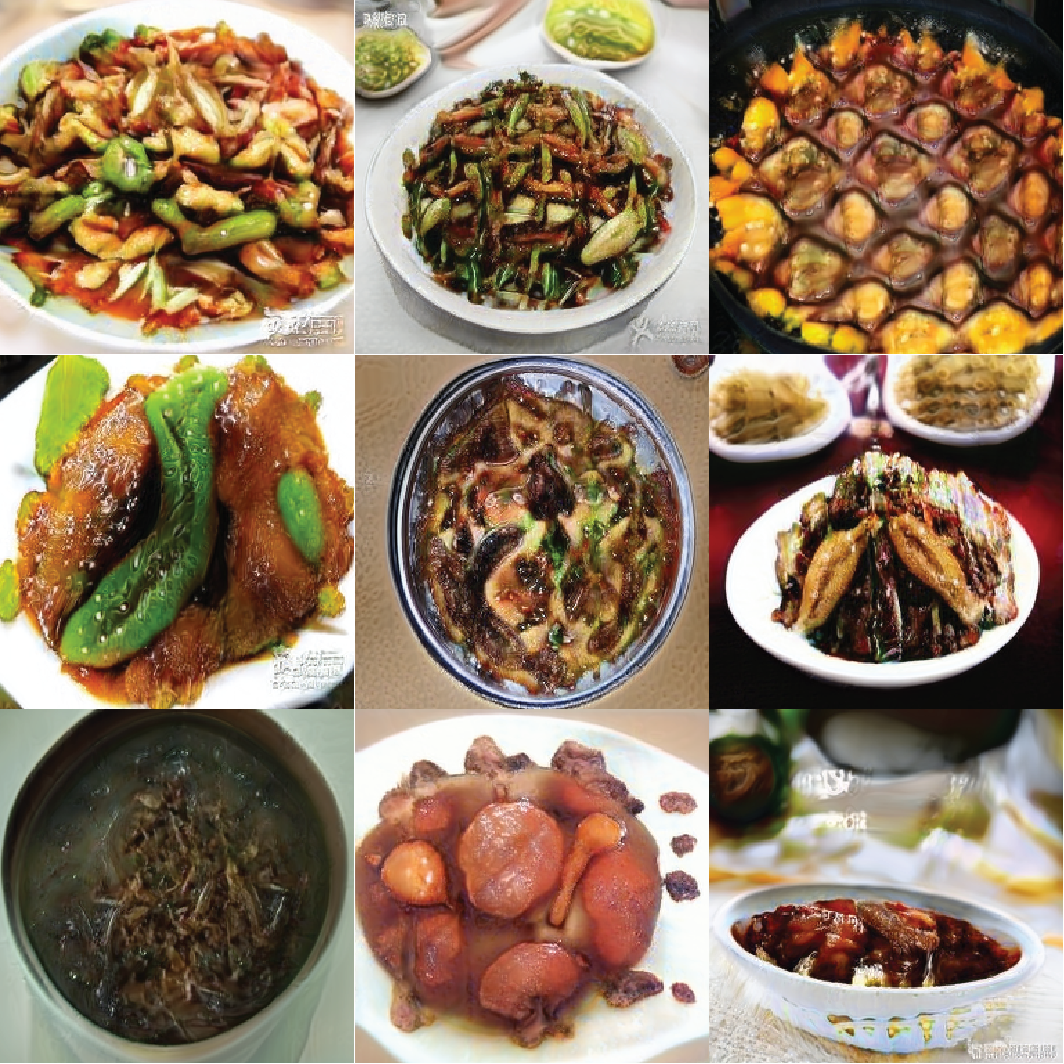}
        \caption{}
        \label{fig:stylegan_good}
    \end{subfigure}
    \begin{subfigure}[t]{0.45\textwidth}
        \centering
        \includegraphics[height=4.5cm]{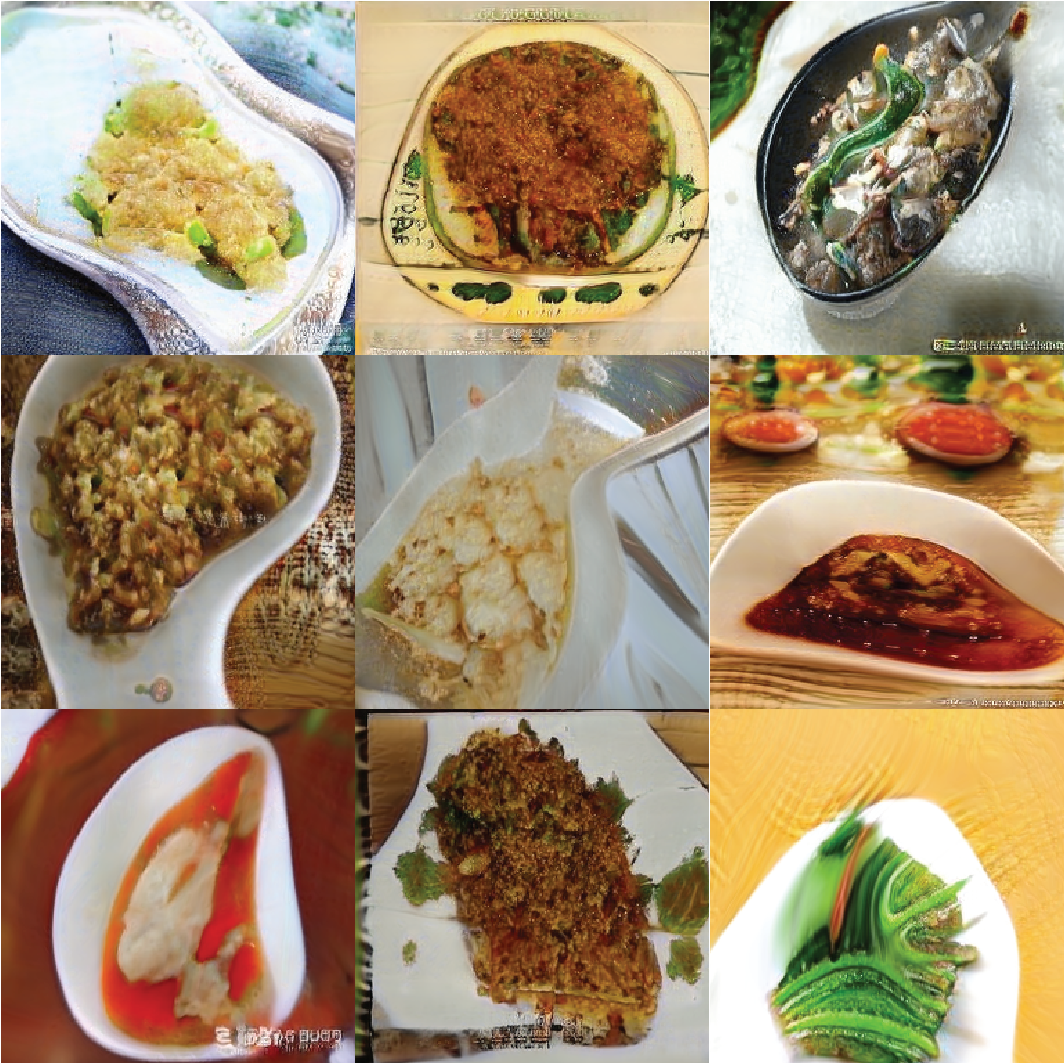}
        \caption{}
        \label{fig:stylegan_bad}
    \end{subfigure}
    \vspace{-0.5em}
    \caption{Image examples generated by StyleGAN3: (a) with round-shaped containers and (b) with irregular-shaped containers.}
    \label{img:stylegan}

\end{figure*}

To calculate the FID values of the generated images, we randomly selected $30,000$ images as input images. For each input image, we generated one output image and calculated the FID value based on the training dataset and the $30,000$ generated images. The result is shown in Table~\ref{table: fid}.
The FID value of our method is $4.97$. 
To benchmark against StyleGAN3\cite{Karras2021}, we ran the StyleGAN3 algorithm using the VireoFood-172 dataset. The model trained $100$ epochs ($10,000$K images), utilizing the default hyper-parameters. Image examples generated by StyleGAN3 are shown in Fig.~\ref{img:stylegan}. We randomly generated $30,000$ images using StyleGAN3 and then calculated the FID value based on these generated images.
The FID value of StyleGAN3 for the VireoFood-172 dataset is $9.25$ as shown in Table~\ref{table: fid}. 

From Fig.~\ref{img:food_gen}, we can see that the structural integrity of the images generated by our network is consistent across different styles. This consistency proves the ability of our method to generate diverse food images while adhering to fixed shape constraints. 
On the contrary, it can be observed from Fig.~\ref{img:stylegan}(b) that the shapes of the containers generated by StyleGAN3 are quite irregular and unpredictable.

The FID value of the $30,000$ generated images when using the Food-101 dataset as the training set is listed in Table~\ref{table: fid_101}. For comparison, the FID values for other models using the same dataset~\cite{Han2023} are also included in this Table. It shows that our model achieves the lowest FID value, $22.82$.

\begin{table}[t]
    \centering
    \begin{tabular}{lc}
        \toprule
        Method & FID\\
        \midrule
        StyleGAN3~\cite{Karras2021} & 9.25 \\ 
        \midrule
        Ours & 4.97 \\
        \bottomrule
    \end{tabular}
    \caption{Comparison of FID on the VireoFood-172 Dataset.}
    \vspace{-0.05in}
    \label{table: fid}
\end{table}

\begin{table}[t]
    \centering
    \begin{tabular}{lc}
        \toprule
        Method & FID\\
        \midrule
        StyleGAN3 \cite{Han2023} 
        & 39.05\\
        Finetuned Latent Diffusion \cite{Han2023}
        & 30.39 \\
        ClusDiff \cite{Han2023} 
        & 27.73 \\ 
        \midrule
        Ours & 22.82\\
        \bottomrule

    \end{tabular}
    \caption{Comparison of FID among various food image generation models on the Food-101 dataset.}
    \vspace{-0.05in}
    \label{table: fid_101}
\end{table}

\begin{figure*}[t]
    \centering
    \includegraphics[height=6.4cm]{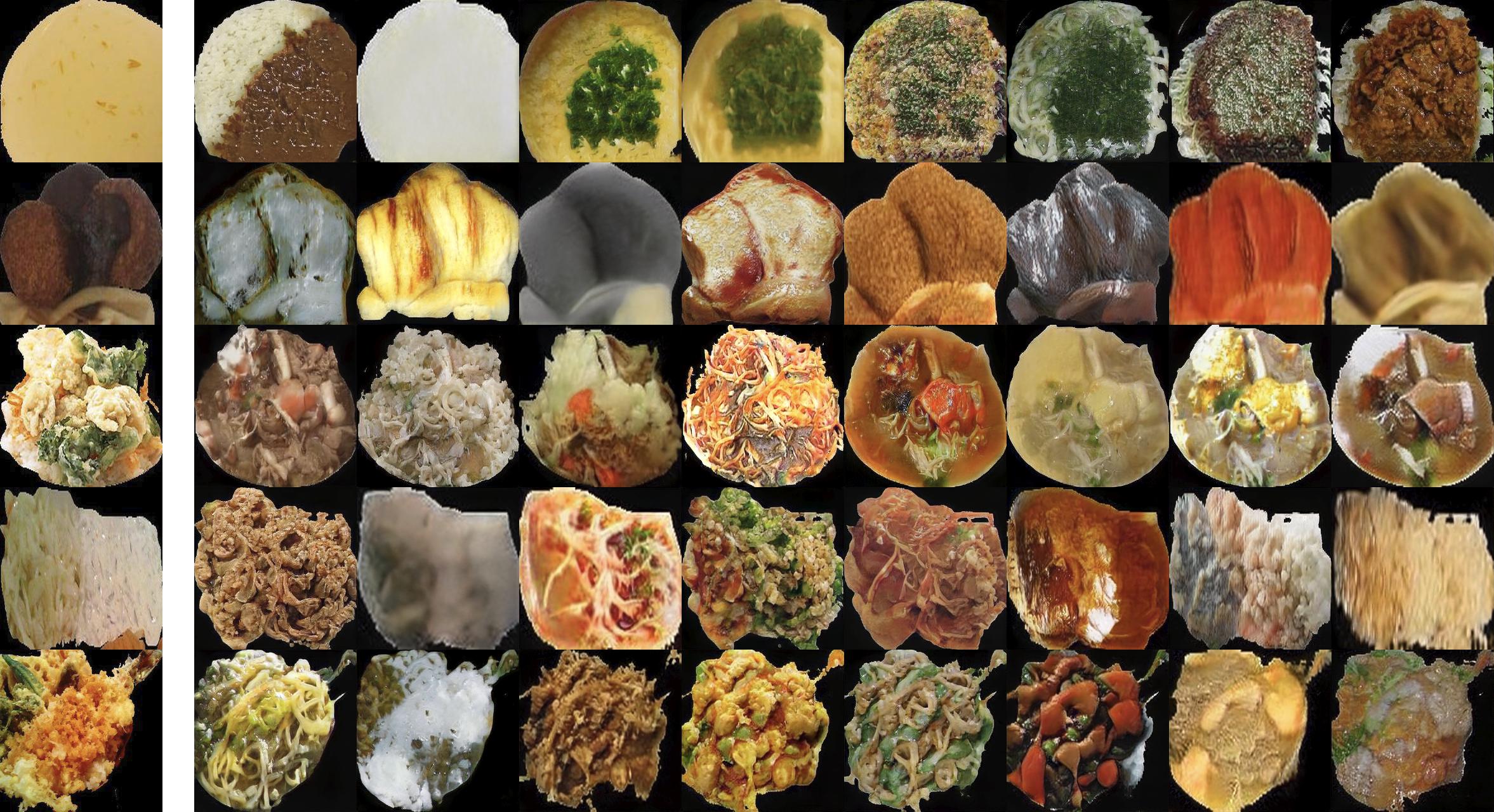}
    \vspace{-0.5em}
    \caption[short]{Image examples generated by our model: The first column is the input images, and the rest are generated images.}
    \label{img:food_gen_seg}
\end{figure*}

\subsection{Evaluation of Shape Preservation Performance }
\label{exp2}

\begin{figure}[tp]
    \centering
    \includegraphics[height=4cm]{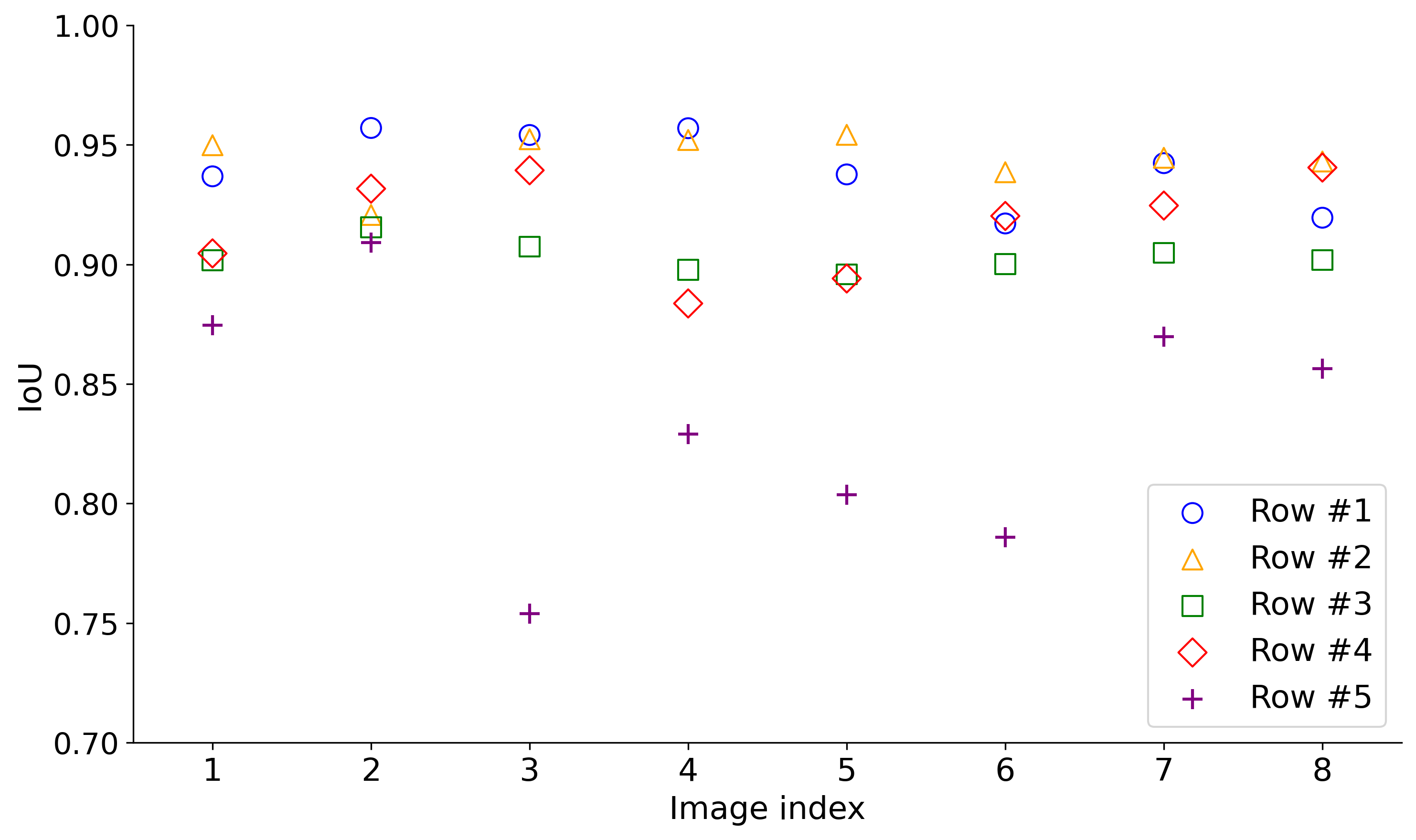}
    \caption[short]{IoU of the generated images shown in Fig.~\ref{img:food_gen_seg}.}
    \label{img:iou}
\end{figure}

In this subsection, we evaluated the shape-preservation performance of the proposed GAN architecture (Fig.~\ref{img:overview}). Here the segmentation images were employed by the training data for the network to learn the texture. The images generated by the network only contain foods. Using segmentation images simplifies the evaluation of the network's performance in maintaining accurate shapes.

\noindent\textbf{Dataset.\ }
We used the segmented food images in the UEC-FoodPIX dataset~\cite{Okamoto2021} for shape-preservation evaluation. This dataset is particularly well-suited for our study as it includes segmentation information for a variety of food items. This allows us to quantitatively assess how well our network preserves the food shape. The UEC-FoodPIX dataset comprises $120$ food classes and a total of $9,000$ images. Each image in the dataset may contain more than one type of food, and the resolutions vary. In pre-processing, we extracted the image of each food item from the original image according to the provided segmentation mask. Then we resized each image to $256 \times 256$ pixel resolution for training.

\noindent\textbf{Evaluation Metric.\ }
Our network is specifically designed to preserve the shape of food in the input image while changing the food category in the generated images. We use the Intersection Over Union (IoU) metric to evaluate how well food shapes are preserved. IoU is a widely used metric in image processing, specifically object detection, for quantifying the degree of overlap between two areas.
To calculate the IoU, we segmented the generated images. A high IoU score signifies effective shape preservation, implying that the network is proficient in replacing the food in the input image with another category of food while maintaining the shape.

\noindent\textbf{Results.\ }
Fig.~\ref{img:food_gen_seg} presents some image examples generated by our network. 
The first column shows the original input images. Subsequent columns display images created by varying the latent variable $z$ while keeping the corresponding input image from the first column fixed.

Fig.~\ref{img:iou} displays the IoU scores achieved by our network. We selected five random images as inputs and generated eight output images for each. We calculated the IoU score for each generated image. In Fig.~\ref{img:iou}, each color corresponds to the IoU scores for the same input image, providing a clear visual representation of the network's performance in terms of shape consistency across multiple outputs. In general, all IoU scores are above $0.8$, and the average IoU score of the eight images is $0.91$.

\begin{figure*}[!ht]
    \centering
    \includegraphics[height=5.5cm]{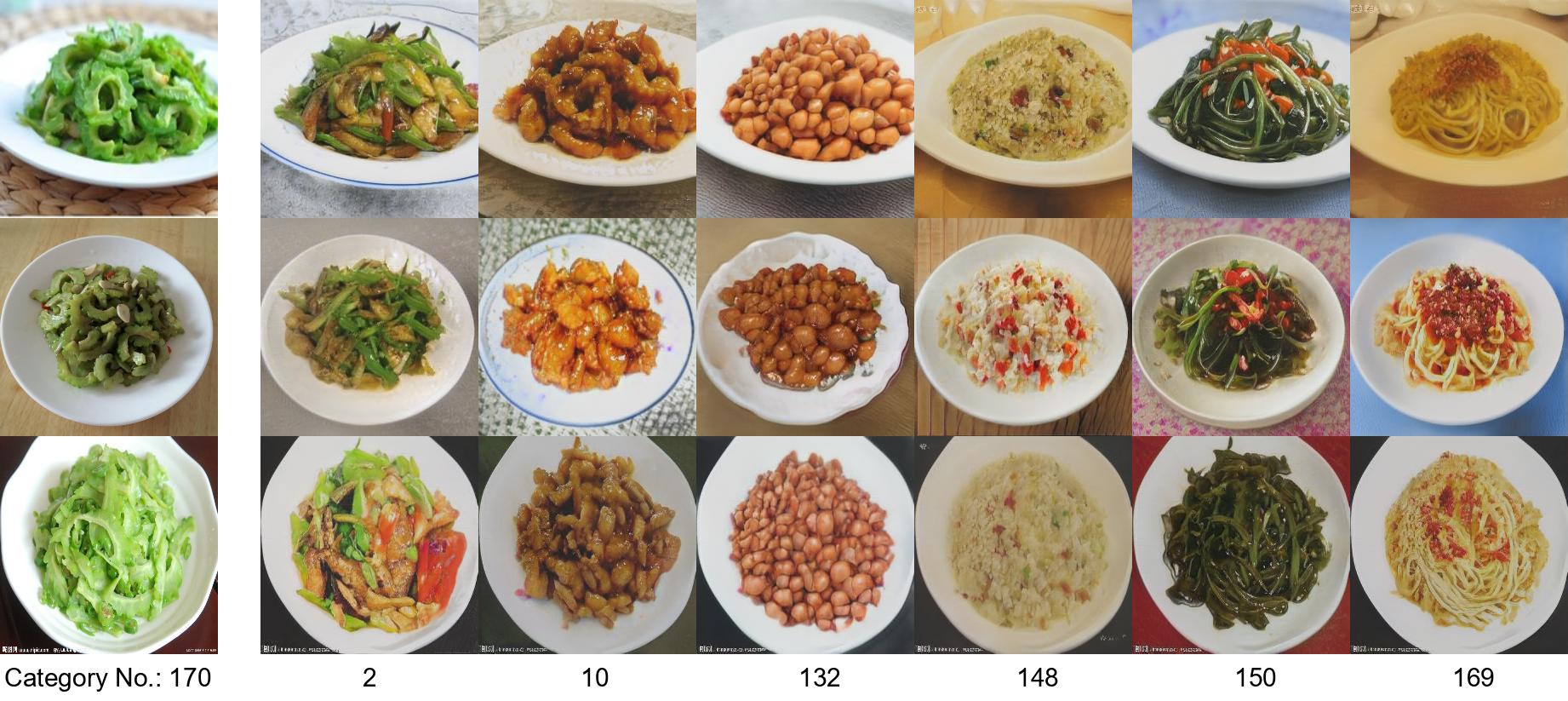}
    \vspace{-0.5em}
    \caption[short]{Image examples generated by our model: The first column is the input images, and the rest are generated by our model. Generated images in each column correspond to the same food category, which is controlled by the category label $c$.}
    \label{img:food_gen_cond}
\end{figure*}

\begin{figure}[!ht]
    \centering
    \includegraphics[height=4cm]{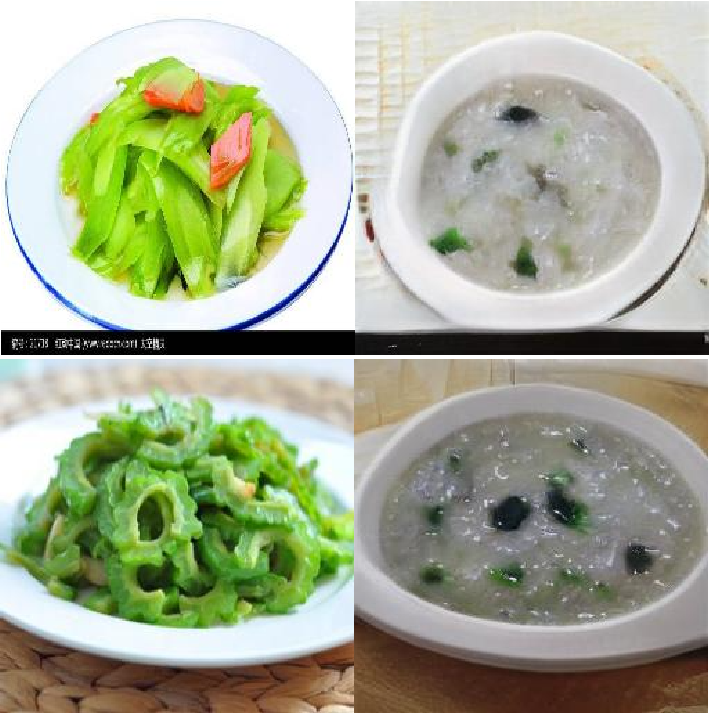}
    \vspace{-0.5em}
    \caption{Image examples when the containers of the input and the output are inconsistent. The first column is the input images, and the second column is the output images, where the fried vegetables are substituted with porridge. The containers for the vegetables (i.e., plate) and the porridge (i.e., bowl) are not matched.}
    \label{fig:discussion3}

\end{figure}
 
\begin{figure}[!ht]
\centering
    \begin{subfigure}[b]{0.43\textwidth}
    \centering
        \includegraphics[height=1.8cm]{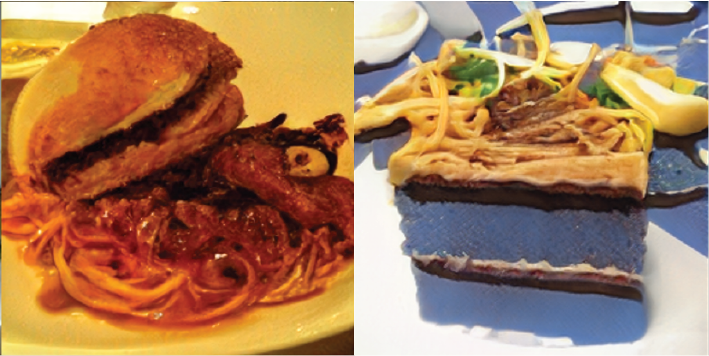}
        \caption{}
        \label{fig:discussion1}
    \end{subfigure}
    \begin{subfigure}[b]{0.43\textwidth}
        \centering
        \includegraphics[height=1.8cm]{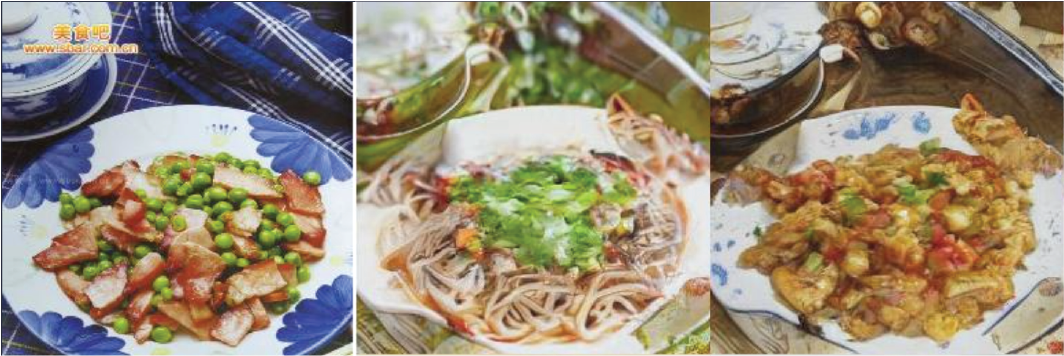}
        \caption{}
        \label{fig:discussion2}
    \end{subfigure}
    \vspace{-0.5em}
    \caption{(a) Image examples of combining incompatible ingredients. (b) Images examples of misinterpreting plate patterns as food. }
    \label{img: discussion}
\end{figure}

\subsection{Category Control of the Generated Images}
\label{exp3}
\begin{table}[t]
    \centering
    \resizebox{\columnwidth}{!}{
    \begin{tabular}{lcccccc}
        \toprule
        Category No. &2&10&132&148 & 150&169\\
        \midrule
        FID &19.57&24.09&50.78&51.12&44.16&39.68\\ 
        \bottomrule
    \end{tabular}
    }
    \caption{FID values of the generated images of six food categories. Image examples of each category are shown in Fig. 7}
    \vspace{-0.05in}
    \label{table: fid_cond}
\end{table}

Generating images without conditional constraints (as in Fig.~\ref{img:food_gen}) can increase the diversity of the generated images, thus they are perfectly suitable for the purpose of volume estimation. However, they are not suitable for image recognition since they do not have labels for food categories. Therefore, to generate images for food recognition, we applied a conditional generator and a conditional discriminator to control the category of the generated images. The output category can be explicitly controlled by variable $c$.

The VireoFood-172 dataset was applied in the experiment. Fig.~\ref{img:food_gen_cond} shows the results. Three random images in one category were selected as the input, which are shown in the first column. Subsequent images in each column are created by different category labels $c$ with the same input image (in the first row) and latent variable $z$. The FID values for one thousand generated images in each category are presented in Table~\ref{table: fid_cond}. The FID values across the whole dataset had also been calculated by generating thirty thousand images with random input images, variable $z$, and category label $c$. This value turns out to be $5.18$. On the other hand, the FID value of each single category is greater than that of the whole dataset, as shown in Table~\ref{table: fid_cond}. It may be caused by the small number of images in these categories, the feature distributions may not be accurately estimated from a small set of data.

By manually selecting the desired image categories for generation, the problem of mismatching between the food and container can be avoided. Fig.~\ref{fig:discussion3} illustrates the generated images when the containers of the input and the output are inconsistent. The input is a plate of fried vegetables, and the output is a bowl of porridge. It is impossible to keep the shape of a plate when transferring between these two kinds of foods. In addition, the volume of the food in the bowl cannot be assumed to be close to the volume of the food on the plate. These issues may be solved by providing a plate as the input and excluding categories where the food is typically served only in bowls from the generation process.

\subsection{Implementation Details}
Our model was implemented by PyTorch. The encoder was self-built based on the structure described in Section 3. In our experiments, the dimensions of images in the dataset were different. To accommodate these differences, the number of blocks in the encoder and generator was adjusted while the resolution of features $f$ was fixed. We used the Adam optimizer to train the network, with the learning rate set to $0.0001$ for the first $100$ epochs and then reduced to $0.00001$ for the remaining $150$ epochs. The generator and discriminator were trained with a batch size of $64$. In Sections~\ref{exp1} and ~\ref{exp2}, the category label $c$ is set to ``None.'' In Section~\ref{exp3}, $c$ is a one-hot vector which encodes the category label. In the experiments, the hyper-parameter $\lambda$ was set to $50$.

\section{Discussions}
In our experiments, the foods in the randomly generated images (i.e., without condition control) sometimes may not correspond to real-world foods. Multiple ingredients were randomly mixed to construct a dish, such as noodles in a burger and chips on a cake, as shown in Fig.~\ref{img: discussion}(a). 
While these “strange” foods can still be used for volume estimation, it is challenging to assign categories to these foods. Occasionally, the decoration pattern on the plate can be mistakenly recognized as food, causing the food shape in the generated images to extend into the plate area, as illustrated in Fig.~\ref{img: discussion}(b).

These issues might be attributed to inaccurate learning of the network. Expanding the dataset with food images containing various cuisines, appearances, and patterned food containers can enhance training. However, image augmentation may be unnecessary when diverse real-world food images are already available. Currently, a practical approach is deliberately selecting shape reference images and controlling generated image categories, though this may limit the diversity of the generated images. Conducting iterative training sessions with feedback from human annotators to continuously fine-tune the model holds the potential for enhancing the network's performance over time.

\section{Conclusion}
Our method can generate new images after training with a given image food dataset. The content and volume/shape of the food can be controlled separately. The textures (food category) of the generated image can be controlled by style variable $z$ or category label $c$, and the shape of the generated image can be constrained by the reference image and the encoder.
The generated images are suitable for training deep networks for food recognition and volume estimation, which overcomes the lack of training data in automated dietary assessment. 
\\

\noindent\textbf{Acknowledgment. \ }
This work was supported in part by the U.S. National Institutes of Health [Grants No. R56 DK113819 and No. R01DK127310] and the Bill \& Melinda Gates
Foundation [OPP1171395].

{
    \small
    \bibliographystyle{ieeenat_fullname}
    \bibliography{reference}

\begin{thebibliography}{74}
\providecommand{\natexlab}[1]{#1}
\providecommand{\url}[1]{\texttt{#1}}
\expandafter\ifx\csname urlstyle\endcsname\relax
  \providecommand{\doi}[1]{doi: #1}\else
  \providecommand{\doi}{doi: \begingroup \urlstyle{rm}\Url}\fi

\bibitem[Amoutzopoulos et~al.(2020)Amoutzopoulos, Page, Roberts, Roe, Cade, Steer, Baker, Hawes, Galloway, Yu, and Almiron-Roig]{Amoutzopoulos2020}
Birdem Amoutzopoulos, Polly Page, Caireen Roberts, Mark Roe, Janet Cade, Toni Steer, Ruby Baker, Tabitha Hawes, Catherine Galloway, Dove Yu, and Eva Almiron-Roig.
\newblock Portion size estimation in dietary assessment: a systematic review of existing tools, their strengths and limitations.
\newblock \emph{Nutrition Reviews}, 78\penalty0 (11):\penalty0 885--900, 2020.

\bibitem[Arslan et~al.(2022)Arslan, Memis, Sonmez, and Batur]{Arslan2022}
Berker Arslan, Sefer Memis, Elena~Battini Sonmez, and Okan~Zafer Batur.
\newblock Fine-grained food classification methods on the uec food-100 database.
\newblock \emph{{IEEE} Transactions on Artificial Intelligence}, 3\penalty0 (2):\penalty0 238--243, 2022.

\bibitem[Baek et~al.(2021)Baek, Choi, Uh, Yoo, and Shim]{Baek2021}
Kyungjune Baek, Yunjey Choi, Youngjung Uh, Jaejun Yoo, and Hyunjung Shim.
\newblock Rethinking the truly unsupervised image-to-image translation.
\newblock In \emph{2021 {IEEE/CVF} International Conference on Computer Vision}. IEEE, 2021.

\bibitem[Baranowski(2012)]{Baranowski2012}
Tom Baranowski.
\newblock \emph{{24-Hour} recall and diet record methods}, pages 49--69.
\newblock Oxford University Press New York, NY, 2012.

\bibitem[Battini~Sönmez et~al.(2023)Battini~Sönmez, Memiş, Arslan, and Batur]{Sonmez2023}
Elena Battini~Sönmez, Sefer Memiş, Berker Arslan, and Okan~Zafer Batur.
\newblock The segmented {UEC} food-100 dataset with benchmark experiment on food detection.
\newblock \emph{Multimedia Systems}, 29\penalty0 (4):\penalty0 2049--2057, 2023.

\bibitem[Bossard et~al.(2014)Bossard, Guillaumin, and Van~Gool]{bossard14}
Lukas Bossard, Matthieu Guillaumin, and Luc Van~Gool.
\newblock Food-101 -- mining discriminative components with random forests.
\newblock In \emph{European Conference on Computer Vision}, pages 446--461. Springer International Publishing, 2014.

\bibitem[Brock et~al.(2018)Brock, Donahue, and Simonyan]{Brock2018}
Andrew Brock, Jeff Donahue, and Karen Simonyan.
\newblock Large scale {GAN} training for high fidelity natural image synthesis.
\newblock \emph{arXiv preprint arXiv:1809.11096}, 2018.

\bibitem[Bynagari(2019)]{Bynagari2019}
Naresh~Babu Bynagari.
\newblock {GANs} trained by a two time-scale update rule converge to a local nash equilibrium.
\newblock \emph{Asian Journal of Applied Science and Engineering}, 8\penalty0 (1):\penalty0 25--34, 2019.

\bibitem[Cecchini et~al.(2010)Cecchini, Sassi, Lauer, Lee, Guajardo-Barron, and Chisholm]{Cecchini2010}
Michele Cecchini, Franco Sassi, Jeremy~A Lauer, Yong~Y Lee, Veronica Guajardo-Barron, and Daniel Chisholm.
\newblock Tackling of unhealthy diets, physical inactivity, and obesity: health effects and cost-effectiveness.
\newblock \emph{The Lancet}, 376\penalty0 (9754):\penalty0 1775--1784, 2010.

\bibitem[Chae et~al.(2011)Chae, Woo, Kim, Maciejewski, Zhu, Delp, Boushey, and Ebert]{Chae2011}
Junghoon Chae, Insoo Woo, SungYe Kim, Ross Maciejewski, Fengqing Zhu, Edward~J. Delp, Carol~J. Boushey, and David~S. Ebert.
\newblock Volume estimation using food specific shape templates in mobile image-based dietary assessment.
\newblock In \emph{{SPIE} Proceedings}. {SPIE}, 2011.

\bibitem[Chen et~al.(2021)Chen, Jia, Zhao, Mao, Lo, Anderson, Frost, Jobarteh, McCrory, Sazonov, Steiner-Asiedu, Ansong, Baranowski, Burke, and Sun]{Chen2021}
Guangzong Chen, Wenyan Jia, Yifan Zhao, Zhi-Hong Mao, Benny Lo, Alex~K. Anderson, Gary Frost, Modou~L. Jobarteh, Megan~A. McCrory, Edward Sazonov, Matilda Steiner-Asiedu, Richard~S. Ansong, Thomas Baranowski, Lora Burke, and Mingui Sun.
\newblock {Food/Non-Food} classification of real-life egocentric images in low- and middle-income countries based on image tagging features.
\newblock \emph{Frontiers in Artificial Intelligence}, 4, 2021.

\bibitem[Chen et~al.(2013)Chen, Jia, Yue, Li, Sun, Fernstrom, and Sun]{Chen2013}
Hsin-Chen Chen, Wenyan Jia, Yaofeng Yue, Zhaoxin Li, Yung-Nien Sun, John~D Fernstrom, and Mingui Sun.
\newblock Model-based measurement of food portion size for image-based dietary assessment using {3D/2D} registration.
\newblock \emph{Measurement Science and Technology}, 24\penalty0 (10):\penalty0 105701, 2013.

\bibitem[Chen and wah Ngo(2016)]{Chen2016a}
Jingjing Chen and Chong wah Ngo.
\newblock Deep-based ingredient recognition for cooking recipe retrieval.
\newblock In \emph{Proceedings of the 24th {ACM} International Conference on Multimedia}. {ACM}, 2016.

\bibitem[Chen et~al.(2012)Chen, Yang, Ho, Wang, Liu, Chang, Yeh, and Ouhyoung]{Chen2012}
Mei-Yun Chen, Yung-Hsiang Yang, Chia-Ju Ho, Shih-Han Wang, Shane-Ming Liu, Eugene Chang, Che-Hua Yeh, and Ming Ouhyoung.
\newblock Automatic chinese food identification and quantity estimation.
\newblock In \emph{Proceedings of SIGGRAPH Asia 2012 Technical Briefs}, New York, NY, USA, 2012. ACM.

\bibitem[Christ et~al.(2017)Christ, Schlecht, Ettlinger, Grün, Heinle, Tatavatry, Ahmadi, Diepold, and Menze]{Christ2017}
Patrick~Ferdinand Christ, Sebastian Schlecht, Florian Ettlinger, Felix Grün, Christoph Heinle, Sunil Tatavatry, Seyed-Ahmad Ahmadi, Klaus Diepold, and Bjoern~H. Menze.
\newblock {Diabetes60} — inferring bread units from food images using fully convolutional neural networks.
\newblock In \emph{Proceedings of 2017 {IEEE} International Conference on Computer Vision Workshops}, pages 1526--1535, 2017.

\bibitem[Ciocca et~al.(2017)Ciocca, Napoletano, and Schettini]{Ciocca2017}
Gianluigi Ciocca, Paolo Napoletano, and Raimondo Schettini.
\newblock Food recognition: a new dataset, experiments, and results.
\newblock \emph{{IEEE} Journal of Biomedical and Health Informatics}, 21\penalty0 (3):\penalty0 588--598, 2017.

\bibitem[Dalakleidi et~al.(2022)Dalakleidi, Papadelli, Kapolos, and Papadimitriou]{Dalakleidi2022}
Kalliopi~V Dalakleidi, Marina Papadelli, Ioannis Kapolos, and Konstantinos Papadimitriou.
\newblock Applying image-based food-recognition systems on dietary assessment: a systematic review.
\newblock \emph{Advances in Nutrition}, 13\penalty0 (6):\penalty0 2590--2619, 2022.

\bibitem[Ege et~al.(2019)Ege, Shimoda, and Yanai]{Ege2019}
Takumi Ege, Wataru Shimoda, and Keiji Yanai.
\newblock A new large-scale food image segmentation dataset and its application to food calorie estimation based on grains of rice.
\newblock In \emph{Proceedings of the 5th International Workshop on Multimedia Assisted Dietary Management}. ACM, 2019.

\bibitem[Fang et~al.(2015)Fang, Liu, Zhu, Delp, and Boushey]{Fang2015}
Shaobo Fang, Chang Liu, Fengqing Zhu, Edward~J. Delp, and Carol~J. Boushey.
\newblock Single-view food portion estimation based on geometric models.
\newblock \emph{ISM}, 2015:\penalty0 385--390, 2015.

\bibitem[Fang et~al.(2019)Fang, Shao, Kerr, Boushey, and Zhu]{Fang2019}
Shaobo Fang, Zeman Shao, Deborah~A Kerr, Carol~J Boushey, and Fengqing Zhu.
\newblock An end-to-end image-based automatic food energy estimation technique based on learned energy distribution images: protocol and methodology.
\newblock \emph{Nutrients}, 11\penalty0 (4):\penalty0 877, 2019.

\bibitem[Forouzanfar and et. al.(2015)]{Forouzanfar2015}
Mohammad~H Forouzanfar and Alexander et. al.
\newblock Global, regional, and national comparative risk assessment of 79 behavioural, environmental and occupational, and metabolic risks or clusters of risks in 188 countries, 1990{\textendash}2013: a systematic analysis for the global burden of disease study 2013.
\newblock \emph{The Lancet}, 386\penalty0 (10010):\penalty0 2287--2323, 2015.

\bibitem[Gibson(2005)]{Gibson2005}
Rosalind~S Gibson.
\newblock \emph{Principles of nutritional assessment}.
\newblock Oxford University PressNew York, NY, 2005.

\bibitem[Goodfellow et~al.(2014)Goodfellow, Pouget-Abadie, Mirza, Xu, Warde-Farley, Ozair, Courville, and Bengio]{Goodfellow2014}
Ian Goodfellow, Jean Pouget-Abadie, Mehdi Mirza, Bing Xu, David Warde-Farley, Sherjil Ozair, Aaron Courville, and Yoshua Bengio.
\newblock Generative {adversarial} {nets}.
\newblock In \emph{Advances in Neural Information Processing Systems}. Curran Associates, Inc., 2014.

\bibitem[Han et~al.(2020)Han, Hao, Guerrero, and Pavlovic]{Han2020a}
Fangda Han, Guoyao Hao, Ricardo Guerrero, and Vladimir Pavlovic.
\newblock {MPG}: a multi-ingredient pizza image generator with conditional {StyleGANs}.
\newblock Technical report, 2020.
\newblock arXiv:2012.02821 [cs] type: article.

\bibitem[Han et~al.(2023)Han, He, Gupta, Delp, and Zhu]{Han2023}
Yue Han, Jiangpeng He, Mridul Gupta, Edward~J. Delp, and Fengqing Zhu.
\newblock Diffusion model with clustering-based conditioning for food image generation.
\newblock In \emph{Proceedings of the 8th International Workshop on Multimedia Assisted Dietary Management}. ACM, 2023.

\bibitem[Hassannejad et~al.(2016)Hassannejad, Matrella, Ciampolini, De~Munari, Mordonini, and Cagnoni]{Hassannejad2016}
Hamid Hassannejad, Guido Matrella, Paolo Ciampolini, Ilaria De~Munari, Monica Mordonini, and Stefano Cagnoni.
\newblock Food image recognition using very deep convolutional networks.
\newblock In \emph{Proceedings of the 2nd International Workshop on Multimedia Assisted Dietary Management}, page 41–49, New York, NY, USA, 2016. ACM.

\bibitem[He et~al.(2014)He, Xu, Khanna, Boushey, and Delp]{He2014}
Ye He, Chang Xu, Nitin Khanna, Carol~J. Boushey, and Edward~J. Delp.
\newblock Analysis of food images: features and classification.
\newblock In \emph{Proceedings of 2014 {IEEE} International Conference on Image Processing}, pages 2744--2748. IEEE, 2014.

\bibitem[Honbu and Yanai(2022)]{Honbu2022}
Yuma Honbu and Keiji Yanai.
\newblock Setmealasyoulike: sketch-based set meal image synthesis with plate annotations.
\newblock In \emph{Proceedings of the 7th International Workshop on Multimedia Assisted Dietary Management on Multimedia Assisted Dietary Management}. ACM, 2022.

\bibitem[Ito et~al.(2018)Ito, Shimoda, and Yanai]{Ito2018}
Yoshifumi Ito, Wataru Shimoda, and Keiji Yanai.
\newblock Food image generation using a large amount of food images with conditional {GAN}: ramen{GAN} and recipe{GAN}.
\newblock In \emph{Proceedings of the Joint Workshop on Multimedia for Cooking and Eating Activities and Multimedia Assisted Dietary Management}. {ACM}, 2018.

\bibitem[Jia et~al.(2014)Jia, Chen, Yue, Li, Fernstrom, Bai, Li, and Sun]{Jia2014}
Wenyan Jia, Hsin-Chen Chen, Yaofeng Yue, Zhaoxin Li, John Fernstrom, Yicheng Bai, Chengliu Li, and Mingui Sun.
\newblock Accuracy of food portion size estimation from digital pictures acquired by a chest-worn camera.
\newblock \emph{Public Health Nutrition}, 17\penalty0 (8):\penalty0 1671--1681, 2014.

\bibitem[Karras et~al.(2021)Karras, Aittala, Laine, Härkönen, Hellsten, Lehtinen, and Aila]{Karras2021}
Tero Karras, Miika Aittala, Samuli Laine, Erik Härkönen, Janne Hellsten, Jaakko Lehtinen, and Timo Aila.
\newblock Alias-free generative adversarial networks.
\newblock \emph{arXiv}, 2021.

\bibitem[Kawano and Yanai(2014{\natexlab{a}})]{Kawano2014a}
Yoshiyuki Kawano and Keiji Yanai.
\newblock \emph{Automatic expansion of a food image dataset leveraging existing categories with domain adaptation}, pages 3--17.
\newblock Springer International Publishing, 2014{\natexlab{a}}.

\bibitem[Kawano and Yanai(2014{\natexlab{b}})]{Kawano2014b}
Yoshiyuki Kawano and Keiji Yanai.
\newblock Offline 1000-class classification on a smartphone.
\newblock In \emph{Proceedings of 2014 {IEEE} Conference on Computer Vision and Pattern Recognition Workshops}. IEEE, 2014{\natexlab{b}}.

\bibitem[Konstantakopoulos et~al.(2024)Konstantakopoulos, Georga, and Fotiadis]{Konstantakopoulos2024}
Fotios~S. Konstantakopoulos, Eleni~I. Georga, and Dimitrios~I. Fotiadis.
\newblock A review of image-based food recognition and volume estimation artificial intelligence systems.
\newblock \emph{{IEEE} Reviews in Biomedical Engineering}, 17:\penalty0 136--152, 2024.

\bibitem[Lieffers and Hanning(2012)]{Lieffers2012}
Jessica R.~L. Lieffers and Rhona~M. Hanning.
\newblock Dietary assessment and self-monitoring: With nutrition applications for mobile devices.
\newblock \emph{Canadian Journal of Dietetic Practice and Research}, 73\penalty0 (3):\penalty0 e253--e260, 2012.

\bibitem[Lo et~al.(2020)Lo, Sun, Qiu, and Lo]{Lo2020}
Frank Po~Wen Lo, Yingnan Sun, Jianing Qiu, and Benny Lo.
\newblock Image-based food classification and volume estimation for dietary assessment: a review.
\newblock \emph{{IEEE} Journal of Biomedical and Health Informatics}, 24\penalty0 (7):\penalty0 1926--1939, 2020.

\bibitem[Luo(2021)]{Luo2021a}
Sanbi Luo.
\newblock A survey on multimodal deep learning for image synthesis: applications, methods, datasets, evaluation metrics, and results comparison.
\newblock In \emph{Proceedings of 2021 the 5th International Conference on Innovation in Artificial Intelligence}. ACM, 2021.

\bibitem[Marin et~al.(2021)Marin, Biswas, Ofli, Hynes, Salvador, Aytar, Weber, and Torralba]{Marin2021}
Javier Marin, Aritro Biswas, Ferda Ofli, Nicholas Hynes, Amaia Salvador, Yusuf Aytar, Ingmar Weber, and Antonio Torralba.
\newblock {Recipe1M}+: a dataset for learning cross-modal embeddings for cooking recipes and food images.
\newblock \emph{{IEEE} Transactions on Pattern Analysis and Machine Intelligence}, 43\penalty0 (1):\penalty0 187--203, 2021.

\bibitem[Markham et~al.(2023)Markham, Chen, Tai, and Wong]{Markham2023}
Olivia Markham, Yuhao Chen, Chi-en~Amy Tai, and Alexander Wong.
\newblock {FoodFusion}: a latent diffusion model for realistic food image generation.
\newblock \emph{arXiv preprint arXiv:2312.03540}, 2023.

\bibitem[Martinel et~al.(2018)Martinel, Foresti, and Micheloni]{Martinel2018}
Niki Martinel, Gian~Luca Foresti, and Christian Micheloni.
\newblock Wide-slice residual networks for food recognition.
\newblock In \emph{Proceedings of 2018 {IEEE} Winter Conference on Applications of Computer Vision}, pages 567--576. IEEE, 2018.

\bibitem[Matsuda et~al.(2012)Matsuda, Hoashi, and Yanai]{Matsuda2012}
Yuji Matsuda, Hajime Hoashi, and Keiji Yanai.
\newblock Recognition of multiple-food images by detecting candidate regions.
\newblock In \emph{Proceedings of 2012 {IEEE} International Conference on Multimedia and Expo}, pages 25--30. IEEE, 2012.

\bibitem[Mescheder et~al.(2018)Mescheder, Geiger, and Nowozin]{Mescheder2018}
Lars Mescheder, Andreas Geiger, and Sebastian Nowozin.
\newblock Which training methods for {GANs} do actually converge?
\newblock In \emph{International Conference on Machine Learning}, pages 3481--3490. PMLR, 2018.

\bibitem[Min et~al.(2019)Min, Liu, Luo, and Jiang]{Min2019}
Weiqing Min, Linhu Liu, Zhengdong Luo, and Shuqiang Jiang.
\newblock Ingredient-guided cascaded multi-attention network for food recognition.
\newblock In \emph{Proceedings of the 27th {ACM} International Conference on Multimedia}, page 1331–1339, New York, NY, USA, 2019. ACM.

\bibitem[Min et~al.(2020)Min, Liu, Wang, Luo, Wei, Wei, and Jiang]{Min2020}
Weiqing Min, Linhu Liu, Zhiling Wang, Zhengdong Luo, Xiaoming Wei, Xiaolin Wei, and Shuqiang Jiang.
\newblock {ISIA} food-500: a dataset for large-scale food recognition via stacked global-local attention network.
\newblock In \emph{Proceedings of the 28th {ACM} International Conference on Multimedia}, page 393–401, New York, NY, USA, 2020. ACM.

\bibitem[Moshfegh et~al.(2008)Moshfegh, Rhodes, Baer, Murayi, Clemens, Rumpler, Paul, Sebastian, Kuczynski, Ingwersen, Staples, and Cleveland]{Moshfegh2008}
Alanna~J Moshfegh, Donna~G Rhodes, David~J Baer, Theophile Murayi, John~C Clemens, William~V Rumpler, David~R Paul, Rhonda~S Sebastian, Kevin~J Kuczynski, Linda~A Ingwersen, Robert~C Staples, and Linda~E Cleveland.
\newblock The {US} department of agriculture automated multiple-pass method reduces bias in the collection of energy intakes.
\newblock \emph{The American Journal of Clinical Nutrition}, 88\penalty0 (2):\penalty0 324--332, 2008.

\bibitem[Myers et~al.(2015)Myers, Johnston, Rathod, Korattikara, Gorban, Silberman, Guadarrama, Papandreou, Huang, and Murphy]{Myers2015}
Austin Myers, Nick Johnston, Vivek Rathod, Anoop Korattikara, Alex Gorban, Nathan Silberman, Sergio Guadarrama, George Papandreou, Jonathan Huang, and Kevin Murphy.
\newblock {Im2Calories}: towards an automated mobile vision food diary.
\newblock In \emph{Proceedings of 2015 {IEEE} International Conference on Computer Vision}. {IEEE}, 2015.

\bibitem[Okamoto and Yanai(2021)]{Okamoto2021}
Kaimu Okamoto and Keiji Yanai.
\newblock \emph{{UEC-FoodPix} complete: a large-scale food image segmentation dataset}, pages 647--659.
\newblock Springer International Publishing, 2021.

\bibitem[Ortega et~al.(2015)Ortega, P{\'e}rez-Rodrigo, and L{\'o}pez-Sobaler]{Ortega2015}
Rosa~M Ortega, Carmen P{\'e}rez-Rodrigo, and Ana~M L{\'o}pez-Sobaler.
\newblock Dietary assessment methods: dietary records.
\newblock \emph{Nutricion Hospitalaria}, 31\penalty0 (3):\penalty0 38--45, 2015.

\bibitem[Pan et~al.(2020)Pan, Dai, Hou, Li, and Sheng]{Pan2020}
Siyuan Pan, Ling Dai, Xuhong Hou, Huating Li, and Bin Sheng.
\newblock {ChefGAN}: food image generation from recipes.
\newblock In \emph{Proceedings of the 28th {ACM} International Conference on Multimedia}, page 4244–4252, New York, NY, USA, 2020. Association for Computing Machinery.

\bibitem[Rajesh et~al.(2022)Rajesh, Raghu, and Sangeetha]{Rajesh2022}
Arnav~A Rajesh, Madhumita Raghu, and J Sangeetha.
\newblock Fast food image recognition using transfer learning.
\newblock In \emph{2022 Fourth International Conference on Cognitive Computing and Information Processing}, pages 1--10. IEEE, 2022.

\bibitem[Raju and Sazonov(2021)]{Raju2021}
Viprav~B. Raju and Edward Sazonov.
\newblock A systematic review of sensor-based methodologies for food portion size estimation.
\newblock \emph{{IEEE} Sensors Journal}, 21\penalty0 (11):\penalty0 12882--12899, 2021.

\bibitem[Schatzkin et~al.(2009)Schatzkin, Subar, Moore, Park, Potischman, Thompson, Leitzmann, Hollenbeck, Morrissey, and Kipnis]{Schatzkin2009}
Arthur Schatzkin, Amy~F. Subar, Steven Moore, Yikyung Park, Nancy Potischman, Frances~E. Thompson, Michael Leitzmann, Albert Hollenbeck, Kerry~Grace Morrissey, and Victor Kipnis.
\newblock Observational epidemiologic studies of nutrition and cancer: the next generation (with better observation).
\newblock \emph{Cancer Epidemiology, Biomarkers \& Prevention}, 18\penalty0 (4):\penalty0 1026--1032, 2009.

\bibitem[Shao et~al.(2023)Shao, Vinod, He, and Zhu]{Shao2023}
Zeman Shao, Gautham Vinod, Jiangpeng He, and Fengqing Zhu.
\newblock An end-to-end food portion estimation framework based on shape reconstruction from monocular image.
\newblock In \emph{Proceedings of 2023 IEEE International Conference on Multimedia and Expo}, pages 942--947. IEEE, 2023.

\bibitem[Shim et~al.(2014)Shim, Oh, and Kim]{Shim2014}
Jee-Seon Shim, Kyungwon Oh, and Hyeon~Chang Kim.
\newblock Dietary assessment methods in epidemiologic studies.
\newblock \emph{Epidemiology and Health}, 36:\penalty0 e2014009, 2014.

\bibitem[Slimani et~al.(2011)Slimani, Casagrande, Nicolas, Freisling, Huybrechts, Ock{\'{e}}, Niekerk, van Rossum, Bellemans, Maeyer, Lafay, Krems, Amiano, Trolle, Geelen, de~Vries, and de~Boer~and]{Slimani2011}
N Slimani, C Casagrande, G Nicolas, H Freisling, I Huybrechts, M~C Ock{\'{e}}, E~M Niekerk, C van Rossum, M Bellemans, M~De Maeyer, L Lafay, C Krems, P Amiano, E Trolle, A Geelen, J~H de Vries, and E~J de Boer~and.
\newblock The standardized computerized 24-h dietary recall method {EPIC} -soft adapted for pan-european dietary monitoring.
\newblock \emph{European Journal of Clinical Nutrition}, 65\penalty0 (S1):\penalty0 S5--S15, 2011.

\bibitem[Smith et~al.(2022)Smith, Adam, Manning, Burrows, Collins, and Rollo]{Smith2022}
Shamus~P. Smith, Marc T.~P. Adam, Grace Manning, Tracy Burrows, Clare Collins, and Megan~E. Rollo.
\newblock Food volume estimation by integrating {3D} image projection and manual wire mesh transformations.
\newblock \emph{{IEEE} Access}, 10:\penalty0 48367--48378, 2022.

\bibitem[Sugiyama and Yanai(2021)]{Sugiyama2021}
Yu Sugiyama and Keiji Yanai.
\newblock Cross-modal recipe embeddings by disentangling recipe contents and dish styles.
\newblock In \emph{Proceedings of the 29th {ACM} International Conference on Multimedia}. ACM, 2021.

\bibitem[Sultana et~al.(2023)Sultana, Ahmed, Masud, Huq, Ali, and Naznin]{Sultana2023}
Jamalia Sultana, Benzir~Md. Ahmed, Mohammad~Mehedy Masud, A.~K.~Obidul Huq, Mohammed~Eunus Ali, and Mahmuda Naznin.
\newblock A study on food value estimation from images: taxonomies, datasets, and techniques.
\newblock \emph{{IEEE} Access}, 11:\penalty0 45910--45935, 2023.

\bibitem[Szegedy et~al.(2015)Szegedy, Liu, Jia, Sermanet, Reed, Anguelov, Erhan, Vanhoucke, and Rabinovich]{Szegedy2015}
Christian Szegedy, Wei Liu, Yangqing Jia, Pierre Sermanet, Scott Reed, Dragomir Anguelov, Dumitru Erhan, Vincent Vanhoucke, and Andrew Rabinovich.
\newblock Going deeper with convolutions.
\newblock In \emph{Proceedings of 2015 {IEEE} Conference on Computer Vision and Pattern Recognition}. IEEE, 2015.

\bibitem[Tahir and Loo(2021)]{Tahir2021}
Ghalib~Ahmed Tahir and Chu~Kiong Loo.
\newblock A comprehensive survey of image-based food recognition and volume estimation methods for dietary assessment.
\newblock \emph{Healthcare (Basel)}, 9\penalty0 (12):\penalty0 1676, 2021.

\bibitem[Tan and Le(2019)]{tan2019efficientnet}
Mingxing Tan and Quoc Le.
\newblock Efficientnet: rethinking model scaling for convolutional neural networks.
\newblock In \emph{International Conference on Machine Learning}, pages 6105--6114. PMLR, 2019.

\bibitem[Tay et~al.(2020)Tay, Kaur, Quek, Lim, and Henry]{Tay2020}
Wesley Tay, Bhupinder Kaur, Rina Quek, Joseph Lim, and Christiani~Jeyakumar Henry.
\newblock Current developments in digital quantitative volume estimation for the optimisation of dietary assessment.
\newblock \emph{Nutrients}, 12\penalty0 (4):\penalty0 1167, 2020.

\bibitem[Thames et~al.(2021)Thames, Karpur, Norris, Xia, Panait, Weyand, and Sim]{Thames2021}
Quin Thames, Arjun Karpur, Wade Norris, Fangting Xia, Liviu Panait, Tobias Weyand, and Jack Sim.
\newblock {Nutrition5k}: towards automatic nutritional understanding of generic food.
\newblock In \emph{Proceedings of the {IEEE/CVF} Conference on Computer Vision and Pattern Recognition}, pages 8903--8911, 2021.

\bibitem[Thompson and Subar(2017)]{Thompson2017}
Frances~E Thompson and Amy~F Subar.
\newblock \emph{Dietary assessment methodology}, pages 5--48.
\newblock Elsevier, 2017.

\bibitem[Wang et~al.(2019)Wang, Sahoo, Liu, Lim, and Hoi]{Wang2019b}
Hao Wang, Doyen Sahoo, Chenghao Liu, Ee-peng Lim, and Steven C.~H. Hoi.
\newblock Learning cross-modal embeddings with adversarial networks for cooking recipes and food images.
\newblock In \emph{Proceedings of 2019 {IEEE/CVF} Conference on Computer Vision and Pattern Recognition}, pages 11564--11573. IEEE, 2019.

\bibitem[Wang et~al.(2020)Wang, Chen, Yang, Bi, and Yu]{Wang2020}
Lei Wang, Wei Chen, Wenjia Yang, Fangming Bi, and Fei~Richard Yu.
\newblock A state-of-the-art review on image synthesis with generative adversarial networks.
\newblock \emph{{IEEE} Access}, 8:\penalty0 63514--63537, 2020.

\bibitem[Wang et~al.(2021)Wang, Laria, van~de Weijer, Lopez-Fuentes, and Raducanu]{Wang2021}
Yaxing Wang, Hector Laria, Joost van~de Weijer, Laura Lopez-Fuentes, and Bogdan Raducanu.
\newblock {TransferI2I}: transfer learning for image-to-image translation from small datasets.
\newblock In \emph{Proceedings of 2021 {IEEE/CVF} International Conference on Computer Vision}, pages 13990--13999. IEEE, 2021.

\bibitem[Willett(2012)]{Willett2012}
Walter Willett.
\newblock \emph{Nutritional epidemiology}.
\newblock Oxford University Press, 2012.

\bibitem[Wu et~al.(2021)Wu, Fu, Liu, Lim, Hoi, and Sun]{Wu2021}
Xiongwei Wu, Xin Fu, Ying Liu, Ee-Peng Lim, Steven~C.H. Hoi, and Qianru Sun.
\newblock A large-scale benchmark for food image segmentation.
\newblock In \emph{Proceedings of the 29th Acm International Conference on Multimedia}, page 506–515, New York, NY, USA, 2021. ACM.

\bibitem[Yanai and Kawano(2015)]{Yanai2015}
Keiji Yanai and Yoshiyuki Kawano.
\newblock Food image recognition using deep convolutional network with pre-training and fine-tuning.
\newblock In \emph{Proceedings of 2015 IEEE International Conference on Multimedia \& Expo Workshops}, pages 1--6, 2015.

\bibitem[Yang et~al.(2021)Yang, Yu, Cao, Xu, Yuan, Zhang, Jia, Mao, and Sun]{Yang2021}
Zhengeng Yang, Hongshan Yu, Shunxin Cao, Qi Xu, Ding Yuan, Hong Zhang, Wenyan Jia, Zhi-Hong Mao, and Mingui Sun.
\newblock Human-mimetic estimation of food volume from a single-view {RGB} image using an {AI} system.
\newblock \emph{Electronics}, 10\penalty0 (13):\penalty0 1556, 2021.

\bibitem[Yeom et~al.(2024)Yeom, Gu, and Lee]{Yeom2023}
Taesun Yeom, Chanhoe Gu, and Minhyeok Lee.
\newblock {DuDGAN}: improving class-conditional {GANs} via dual-diffusion.
\newblock \emph{{IEEE} Access}, pages 1--1, 2024.

\bibitem[Zhao et~al.(2020)Zhao, Yap, Kot, and Duan]{Zhao2020}
Heng Zhao, Kim-Hui Yap, Alex~Chichung Kot, and Lingyu Duan.
\newblock {JDNet}: a joint-learning distilled network for mobile visual food recognition.
\newblock \emph{{IEEE} Journal of Selected Topics in Signal Processing}, 14\penalty0 (4):\penalty0 665--675, 2020.

\bibitem[Zhu and Ngo(2020)]{Zhu2020}
Bin Zhu and Chong-Wah Ngo.
\newblock {CookGAN}: Causality based text-to-image synthesis.
\newblock In \emph{Proceedings of 2020 {IEEE/CVF} Conference on Computer Vision and Pattern Recognition}. {IEEE}, 2020.

\end{thebibliography}
}


\end{document}